\documentclass{article}

\UseRawInputEncoding

\usepackage[affil-it]{authblk}
\usepackage[cmex10]{amsmath} 
\usepackage{relsize}
\usepackage{graphicx}
\usepackage{multiobjective,cite,fixltx2e,balance}
\usepackage{dcolumn}
\usepackage{rotating}

\usepackage{subcaption}% http://ctan.org/pkg/subcaption
\usepackage{tikz}

\usepackage{algpseudocode}
\usepackage{graphicx}
\usepackage{ctable}
\usepackage{lipsum}
\usepackage[comma, sort, authoryear]{natbib}

\usepackage{url}

\begin{document}

\title{Agent-based Simulation with Netlogo to Evaluate AmI Scenarios}

\author{Javier Carbo}
%\thanks{Electronic address: \texttt{javier.carbo@uc3m.es}}
\affil{Department of Informatics, Universidad Carlos III de Madrid,\\ Colmenarejo, Madrid, Spain. javier.carbo@uc3m.es}

\author{Nayat Sanchez-Pi*}
%\thanks{Electronic address: \texttt{nayat@ime.uerj.br}; Corresponding author}
\affil{Computer Science Department, Institute of Mathematics and Statistics, Universidade do Estado do Rio de Janeiro\\Rio de Janeiro (RJ) Brazil. *corr: nayat@ime.uerj.br}

\author{Jos\'e Manuel Molina}
%\thanks{Electronic address: \texttt{molina@ic.uc3m.es}}
\affil{Department of Informatics, Universidad Carlos III de Madrid,\\Colmenarejo, Madrid, Spain. molina@ia.u3m.es}

\date{Dated: \today}

\maketitle

\begin{abstract}    

In this paper an agent-based simulation is developed in order to evaluate an AmI scenario based on agents. Many AmI applications are implemented through agents but they are not compared to any other existing alternative in order to evaluate the relative benefits of using them. The proposal simulation environment developed in Netlogo analyse such benefits using two evaluation criteria: First, measuring agent satisfaction of different types of desires along the execution. Second, measuring time savings obtained through a correct use of context information.
 
So, here, a previously suggested agent architecture, an ontology and a 12-steps protocol to provide AmI services in airports, is evaluated using a NetLogo simulation environment. The present work uses a NetLogo model considering scalability problems of this application domain but using FIPA and BDI extensions to be coherent with our previous works and our previous JADE implementation of them. 
 
The NetLogo model presented simulates an airport with agent users passing through several zones located in a specific order in a map: passport controls, check-in counters of airline companies, boarding gates, different types of shopping. Although initial data in simulations are generated randomly, and the model is just an approximation of real-world airports, the definition of this case of use of Ambient Intelligence through NetLogo agents opens an interesting way to evaluate the benefits of using Ambient Intelligence, which is a significant contribution to the final development of them.

\end{abstract}

\section{Introduction}

Virtual simulations frameworks, such as \citep{mercedesgarijo}, have been widely used to evaluate evacuation plans in indoor 
environments. But there are also other scenarios that are very complex to evaluate and this is the case of Internet of Things (IoT) scenarios. Electronic sensors, which act as autonomous compu- tational devices (smartphones, cameras, i-watches, thermical, infrared sensors, drones, etc.), are rapidly becoming ubiquitous capturing daily life activities in all kinds of environments (at home, at the office, and even on a larger scale,  such as in the so-called smart cities). The ubiquity of sensors makes possible the idea envisioned by Weiser in 1991, which presents a world where computers are embedded in everyday life where people could communicate with these devices providing customized services in a way where the network infrastructure would be transparent to the user itself \citep{weiser1991}. This idea is mostly known as Ambient Intelligence (in advance AmI). AmI emphasizes greater user- friendliness, more efficient service support, user-empowerment, and support for human interactions. In this vision, people are surrounded by intelligent and intuitive software entities embedded in everyday sensors around us, recognizing and responding to the particular needs of individuals in an invisible way \citep{kovacs2006some}. 

AmI represents, in other words, a new generation of user-centered computing environments aiming to find new ways to obtain a better integration of the information technology in everyday life activities obtained by ubiquitous sensors. Ideally, people in an AmI environment do not notice these sensors, but they will benefit from the services they are able to provide. Such sensors are aware of people’s presence in those environments and react to their gestures, actions and context \citep{aarts2001ambient}. AmI environments are then integrated by several autonomous computational devices of modern life ranging from con- sumer electronics to mobile phones. AmI has several spheres of application such as: transportation (for instance providing adaptive bus routes or adaptive traffic lights), health (predicting heart attacks, providing faster ambulance calls, etc),home (providing more efficient energy-uses), etc. Recently the interest in AmI Environments has also been focused on demanding highly innovative services in critical areas such as airports and train stations in order to increase security, reduce the length of lines and to better provide updated travel information.

In order to work efficiently, software running on these sensors may have some knowledge about the user. This means that they need to cooperate with other sensors sharing knowledge about the user without interfering with user’s daily life activities. Due to the highly dynamic properties of the above- mentioned environments, the software system running on sensors faces problems such as: user mobility, service failure, resources and  goal changes which may happen in any moment. To cope with these problems, this system must sense the environment, and act on it over time in pursuit of its own benefit.

That is why there is a need  for a special kind of software that should combine ubiquity, context-awareness, intelligence and natural interaction in an AmI environment. The system has also to adapt not only to changes in the environment to be autonomous and self-managed but also adapt to user requirements and needs. The kind of software that meets such requirements is Agent technology. Agents aim to reproduce human behaviour through abilities such as autonomy, proactivity, adaptability, planning, and so on \citep{WooldridgeyJennings1995}. Agents adapt not only to changes in the environment; to be autonomous and self-managed they also adapt to user requirements and needs. This is the underlying foundation of the concept of agent: computer systems capable of independent actions on behalf of their users \citep{Rosenchein1994}. 

We have been working in the confluence of both research areas for many years. Specifically we have developed a distributed agent-based platform to provide AmI services to users in an airport domain\citep{sanchez-jade},\citep{sancheziberagents}. But we found out that evaluation of AmI systems is a difficult problem and seldom tackled in literature because of the privacy issues, hardware costs and the open and dynamic nature of this kind of system. Then, instead of universal real-life evaluations, the most popular way to evaluate them is to observe their performance in particular application scenarios through virtual simulations like\citep{mercedesgarijo} in which a complex and complete framework to evaluate emergency plans in indoor environments is defined. This is also our case.

This paper presents a two-fold criteria evaluation of the benefits of using AmI in a particular domain application we have previously worked in: an airport. We use NetLogo to simulate particular and collective behaviors in an airport. This NetLogo simulation has the objective of comparing user satisfaction due to the delays of agents in rows with and without AmI. We try to find out how AmI could help when a high number of agents are accessing different services through rows, and through the use of location indications as happens in real-life airports. Specifically, we use simulations to compare extra time-savings and the level of satisfaction of agent’s goals when provided with AmI and without it. Such goals are ,for instance, avoiding missing a plane (this provides major satisfaction), meeting shopping interests (this provides minor satisfaction) and reducing time spent waiting in line (this also provides minor satisfaction).
The rest of the paper is structured as followed: section 2 presents contextualized related work; section 3 is where we summarize our previously defined ontology, protocols, agent architecture and airport scenario; section 4 describes the main contribution of this paper: the coherent adaptation of the elements presented in section 3 to a NetLogo model and the simulation experiments results. At last, conclusions are presented.

\section{Foundations}

In the literature, there are several approaches to developing platforms, frameworks and applications for offering context-aware services where agent technology has been applied (as we do) in order to provide the right information at the right time to its users. These applications also include location-based services as our work uses Aruba technology to perform such a  task) providing information and events to the user \citep{crumpet}. 

Application domains of this combination of the three elements: AmI, Agents and Location Technology, they are: TeleCARE project for supporting virtual elderly assistance communities \citep{Afsarmanesh03}; a planning agent AGALZ using case- based reasoning to respond to events and to monitor Alzheimer patients’ health care in execution time \citep{corchado08}; SMAUG \citep{nieto04} is a multi-agent context-aware system that allows tutors and pupils of a university to fully manage their activities; AmbieAgents \citep{lech04} proposes an agent-based infrastructure for context-based information delivery for mobile users; there is also a case study that consists of solving the automation of the internal mail management of a department that is physically distributed in a single floor of a building plant (a restricted and well-known test environment) using ARTIS agent architecture \cite{bajo08}; and an AmI architecture to provide an agent-based surveillance system applying an agent-orientated methodology \citep{pavon07}. None of them, however, has been applied to an airport domain as we have been doing during the past years. 

On the other hand, as agents seem to be the appropriate solution for AmI environments since they provide autonomy and proactivity. In \citep{OHare2004}, O’Hare et al. advocate the use of agents (as do we) as  key enablers in the delivery of AmI. It could be assumed that agents are abstractions for the interaction within an AmI environment, and the single aspect that agents need to ensure is that their behaviour is coordinated. This assumption leads to the use of very simple reactive agents without any cognitive capability \citep{brooks}. But depending on the domain, agents reproducing intelligent behaviours need decision rules that take into consideration complex context information (location, user profile, type of device, etc.) in which these interactions take place  and which has to be interpreted. Complex knowledge processing is required in order to offer, provide and consume services on behalf of humans. We need agents to help humans in their knowledge-related tasks; agents that can somehow understand people’s emotions and rational behaviour, or that can at least attempt to process complex information on our behalf. In this regard, the so called cognitive architecture accomplishes not only the task of regulating the interaction but also of managing complex decision-making. The most extended and promising cognitive architecture is based on the Belief-Desire-Intention paradigm. These three levels of knowledge allow agents to cope with complex decisions supposedly as humans do, following a particular reasoning algorithm \citep{rao}. Communications between agents also attempt to emulate human dialogs through the use of predefined sequences of linguistic performatives as IEEE-accepted FIPA communication standards define them \citep{FIPA}.

BDI-based agent platforms such as JADEX \citep{pokahr+03jadex} or JASON \citep{jason}, and FIPA-compliant platforms such as JADE \citep{JADE} produce agents that are often conceptually heavy models and intensive CPU consuming implementations. This makes them difficult to use as simulation tools when a relatively high number of agents are involved, as we observed with our JADE implementation of the airport AmI application. 

Therefore, an alternative is the use of the lightweight-agent paradigm extended in the simulation research area, known as Multi-Agent based Social Simulation MABS, which is largely used in economics, traffic flow, etc. It allows analysis of complex interactions with heterogeneous individuals \citep{Sichman1998}, and typically represents agents in a very simplistic or atomistic approach. This simplification is needed to avoid the complexity of BDI-based, FIPA-compliant agents. These kinds of simple agents are produced by platforms such  as MASON (cs.gmu.edu/eclab/projects/mason), RePast (repast.sourceforge.net), SMNP (www.monfox.com/dsnmp sim.html) and Netlogo (ccl.northwestern.edu/
netlogo). Some approaches try to address this limitation through the inclusion of cognitive skills in MABS platforms as \citep{botia} did with the integration of MASON and JASON, and like the proposed FIPA and BDI extensions \citep{netlogobdi}to NetLogo do in our work.

Specifically, NetLogo is a programmable modeling environment for simulating natural and social phenomena. It is particularly well suited for modeling complex systems that develop over time. Developers can give instructions to hundreds or thousands of independent agents all operating concurrently. This makes it possible to explore the connection between the micro-level behavior of individuals and the collective behavior that emerge from the interaction of many individuals. Two approaches very close to ours which  use NelLogo models to aid crowd evacuation in emergency situations are \citep{wagner2014agent} and \citep{dawson}.

A wide variety of computational approaches have been proposed for simulation of collective behavior \citep{Pan2007}. In this work, the authors define three classification categories (1) fluid or particle systems, (2) matrix-based systems, and (3) emergent systems, but there are also specific AmI simulators designed to evaluate general AmI systems such as UbiWise \citep{ubiwise}, Tatus \citep{tatus} and UbiReal \citep{ubireal}. However, they are focused on the interaction of a real user with the system and are not designed to develop and run fully automated executions of a particular AmI scenario as we do.

\section{Problem definition}

In this work, the scenario is defined by a 2D grid of pixels, where special rooms are represented by a pixel accessible from any neighbour pixel. Each individual is represented by an autonomous entity, an agent, whose main goals are either taking the plane or recovering baggage and leaving the airport . Many individuals can be located in the same pixel but in each iteration just one of them is interacting with the services/information provided by the room. Several agents are defined with cognitive capacity based on BDI model. This means that each agent has a set of beliefs that include the relevant locations (pixels) the agent has visited, personal beliefs about itself, and beliefs based on information/services received from other agents. This belief sets changes while it moves through the grid and when new information/services arrive from other agents. 
 
\subsection{AmI in an airport}

AmI has applications for different sectors in daily life. One important sector is transportation, specifically airports. AmI intelligence can be presented in this domain as an information system to offer customized services to different types of users (agent roles): passengers, crew and airline staff. We are familiar with this specific problem because we have been working for  years with this application domain \citep{sanchez07}. We previously developed a central- ized system using Appear Networks Platform (www.appearnetworks.com) and Aruba Wi-Fi Location System (www.arubanetworks.com) and later we devel- oped a distributed agent-based platform using the same technology \citep{sanchezjade}, \citep{sancheziberagents}. In both approaches, we assume an initial minimal known profile of the user: name (identifier), agent role, passport data (nationality, physical aspect), suitcases carried, shopping interests and travel info (flight numbers, companies, origin and destination) in order to suggest the best-fitting services. 

Knowledge involved to provide context-awareness in an airport was also defined in an ontology. To build the ontology, we have followed Noy and McGuiness’s proposal which consists in an iterative process based on the methodology proposed by Gruninger and Fox \citep{gruninger1995methodology} who defined the competency questions used in the scope and goal step, and the development of the classes hierarchy based on Top-Down and Bottom-Up strategies.  In our previous works \citep{DBLP:conf/ideal/FuentesCM06}, we defined the problem of context definition in ubiquitous applications. The high level ontology definition that we have described follows the categorization defined by Schilit \citep{schilit1994}, which divided contextual information into a computing context (network, devices, etc.), user context (preferences, location etc.) and physical context (temperature, traffic, etc.). The ontology definition gathers these concepts and their properties and relationships to accomplish this contextual definition.  Important contextual information about the user to take into account is the lo- cation. In order to acquire location information we use Aruba Networks which is a location tracking solution that uses an enterprise-wide WLAN deployment to provide precise location tracking of any Wi-Fi device in the research facility. The RF Locate application can track and locate any Wi-Fi device within range of the Aruba mobility infrastructure. Using accurate deployment,  layouts and triangulation algorithms devices can be easily located including PDAs, rogue APs/Clients, VoWLAN phones, laptops, Wi-Fi asset management tags. Al- though many alternatives exist, most successful indoor location techniques are based on the RSSI triangulation method. But basic RSSI triangulation does not provide sufficient accuracy of location information for many of the users. While techniques such as analysis of building material and walk around calibration can improve the accuracy of RSSI measurements, they add considerable expense and complexity to the network installation. Furthermore,  the improvement in accuracy erodes over time, as the environment changes. WLANs are cellular, where neighbouring APs operate on different RF frequencies (channels) to avoid interference. The Wi-Fi medium access control layer (MAC) allows any station in a basic service set to transmit at any time. Therefore all stations (including the AP) should be listening on the cells RF channel all the time, to avoid miss- ing transmissions. The aforementioned explained the use of time-stealing APs to monitor other channels while nominally providing coverage of their own cell. An alternative technique is to deploy dedicated RF monitors called Air Monitors (AMs). Such monitors are identical to APs (the same hardware and software), but they are configured permanently in the listening mode. This is a very useful capability, because the AMs contribute not only to location accuracy but they also improve security coverage by detecting RF sources that may be security risks or interferers. The drawback of using dedicated AMs is that they add to the capital costs of the network. 
When a wireless device enters the network, the position of the client device is immediately established . Once the client is localized, he can negotiate the set of applications depending on his physical position. In our distributed approach based on agents, entities are in charge of distributing contextual information in order to access the information in a more efficient way. 

\subsection{Agent System architecture}

\begin{figure*}[!t]
\begin{center}
\includegraphics[width=\textwidth]{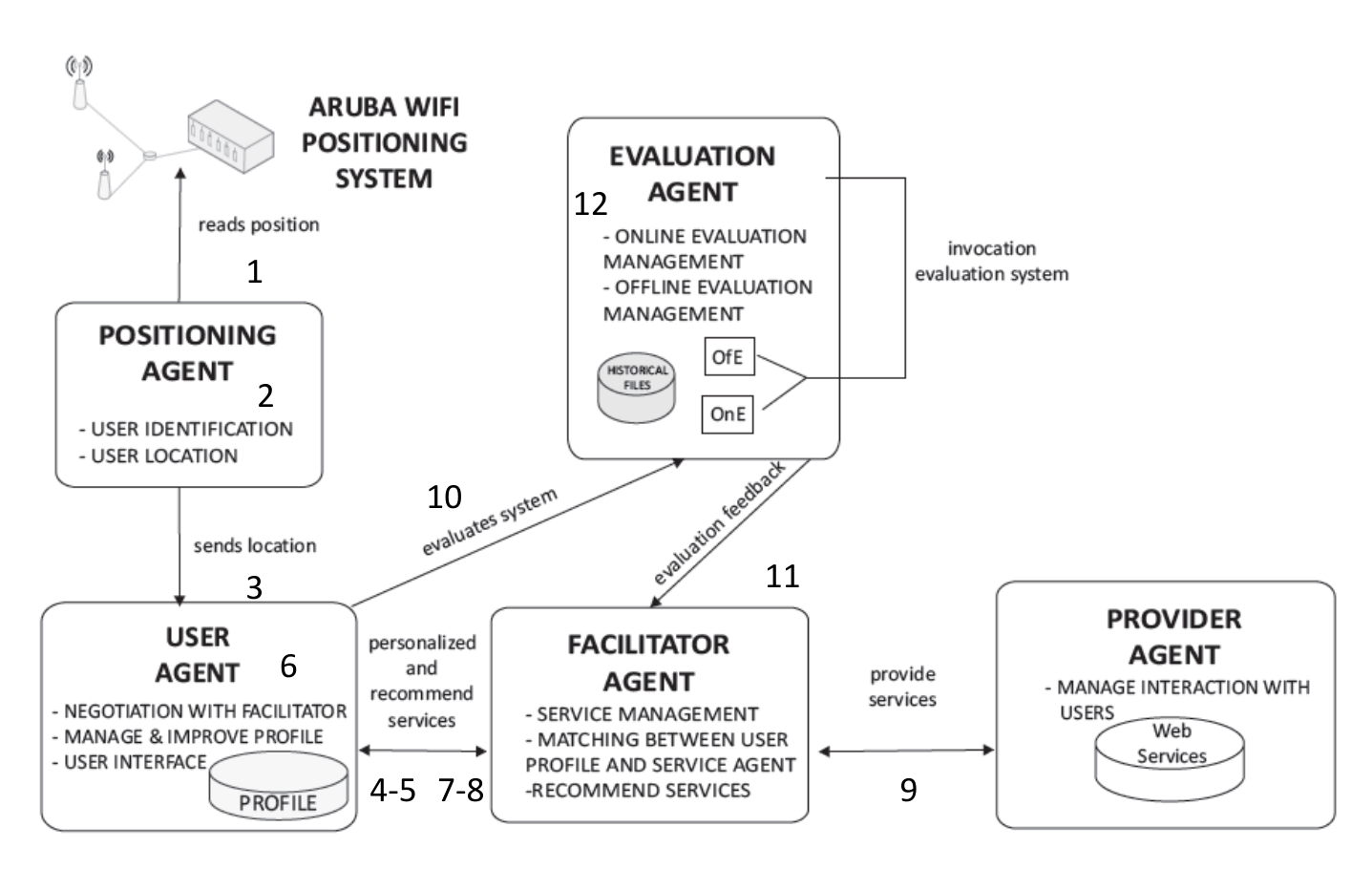}
\caption{Schema of the multi-agent architecture.}
\label{fig:archi}
\end{center}
\end{figure*}

The proposed agent-based architecture manages context information to provide personalized services to users. As it can be observed in Figure \ref{fig:archi}, it consists of five different types of agents that cooperate to provide an adapted service. \textit{User agents} are configured into mobile devices or PDAs. \textit{Provider Agents} supply the different services in the system. A \textit{Facilitator Agent} links the different positions to the providers and services defined in the system. A \textit{Positioning Agent} communicates with the Aruba positioning system \citep{sanchez07} to extract and transmit positioning information to other agents in the system. Finally, an \textit{Evaluator Agent} stores log file in order to acquire a future evaluation criteria of the MAS system developed for AmI scenarios.

Eight concepts have been defined for the ontology of the system. The definition is: \textit{Position } (XCoordinate int,	YCoordinate int), \textit{Place} (Building int,	Floor int), Service (Name String), \textit{Product} (Name String, Characteristics: List of Feature), \textit{Feature} (Name String,	Value String), \textit{Context} (Name String, Characteristics: List of Features), \textit{Profile} (Name: String,	Characteristics: List of Features). Our ontology also includes six predicates with the following arguments:

Our ontology also includes five predicates and an action with the following arguments: \textit{HasLocation} (place, Position, AID), \textit{HasServices} (Place, Position, List of Services), \textit{isProvider} (Place, Position, AID, Service), \textit{HasContext} (What, Who), HasProfile (Profile, AID), and \textit{Provide} (Product, AID).

The interaction with the different agents follows a process which is comprised of the following phases:
\begin{enumerate}
\item The ARUBA positioning system is used to extract information about the positions of the different agents in the system. This way, it is possible to know the positions of the different User Agents and thus extract information about the different Providers Agents that are available for this location.
\item The Positioning Agent reads the information about position (coordinates x and y) and place (building and floor) provided by the Aruba Positioning Agent by reading it from a file, or by processing manually introduced data.
\item The Positioning Agent (Positioning Agent.Send\\Location) communicates the position and place information to the User Agent.
\item Once a User Agent is aware of its own location, it communicates this information to the Facilitator Agent in order to find out the different services available in that location.
\item The Facilitator Agent informs the User Agent about the services available in this position.
\item The User Agent decides the services in which it is interested.
\item Once the User Agent has selected a specific service, it communicates its decision to the Facilitator Agent and queries it about the service providers that are available.
\item The Facilitator Agent informs the User Agent about the identifier of the Provider Agent that supplies the required service in the current location.
\item The User Agent asks the Provider Agent for the required service through the Facilitator Agent.
\item Once the interaction with the Provider Agent is finished, the User Agent provides the evaluation information to the Evaluator Agent.
\item The Evaluator Agent updates the contents of the user profile  with the evaluation information and sends this information to the Evaluator Agent.
\item The Evaluator Agent stores this user profile for future further analysis.
\end{enumerate}

The corresponding number of each phase is shown in Figure \ref{fig:archi} to facilitate the understanding of the communication flow between agents to request a particular service. So with this agent definition, ontology and protocols, we have completely defined the AmI agent-based application domain which we will evaluate using a simulated model of an airport.

\section{Agent simulation with NetLogo}

\begin{figure}
 \centering
  \subfloat[Ingoing 1]{
   \label{fig:perceptionzingoingami}
    \includegraphics[height=7.5 cm]{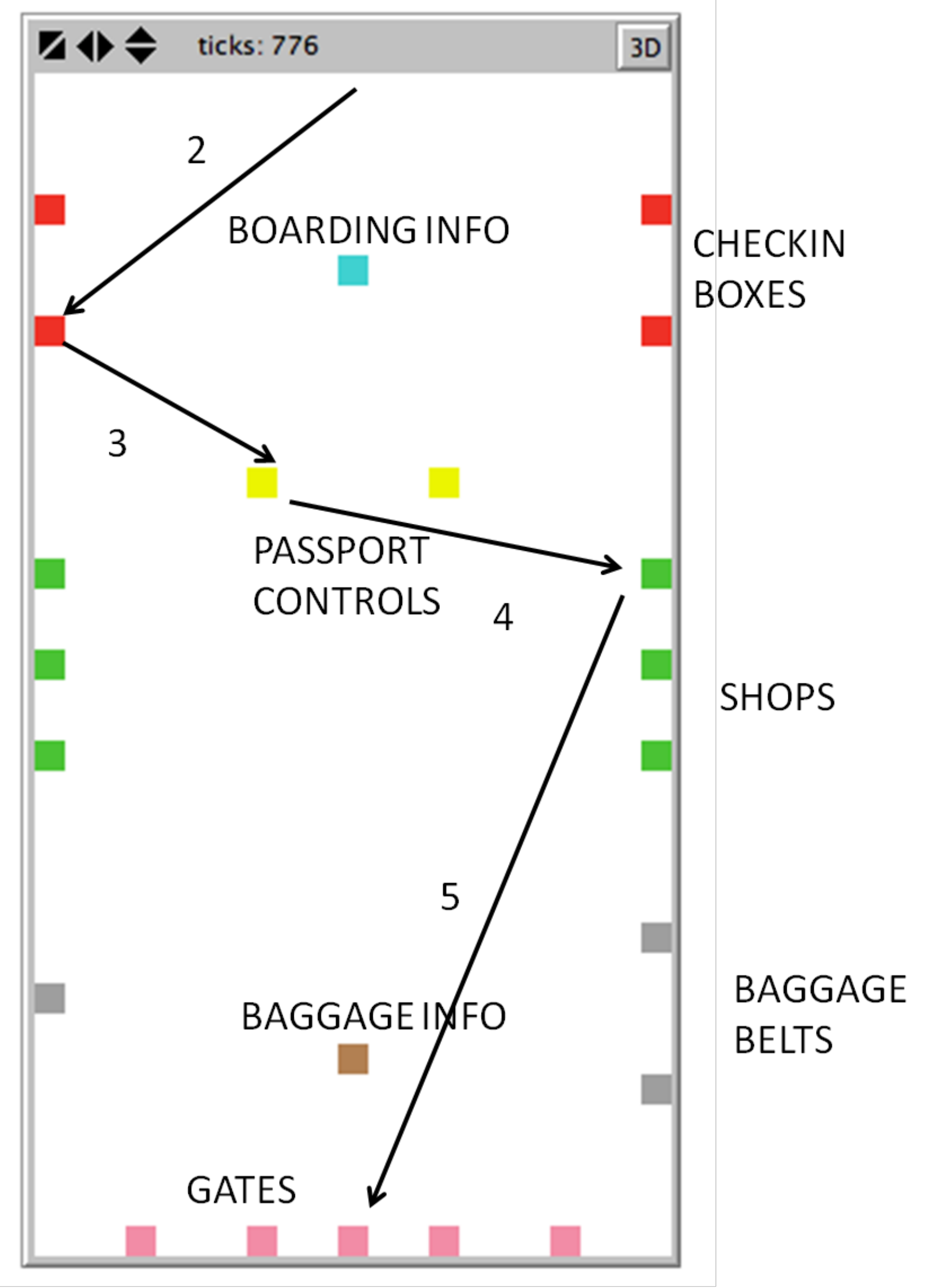}}
  \subfloat[Ingoing 2]{
   \label{fig:perceptionzingoing}
    \includegraphics[height=7.5 cm]{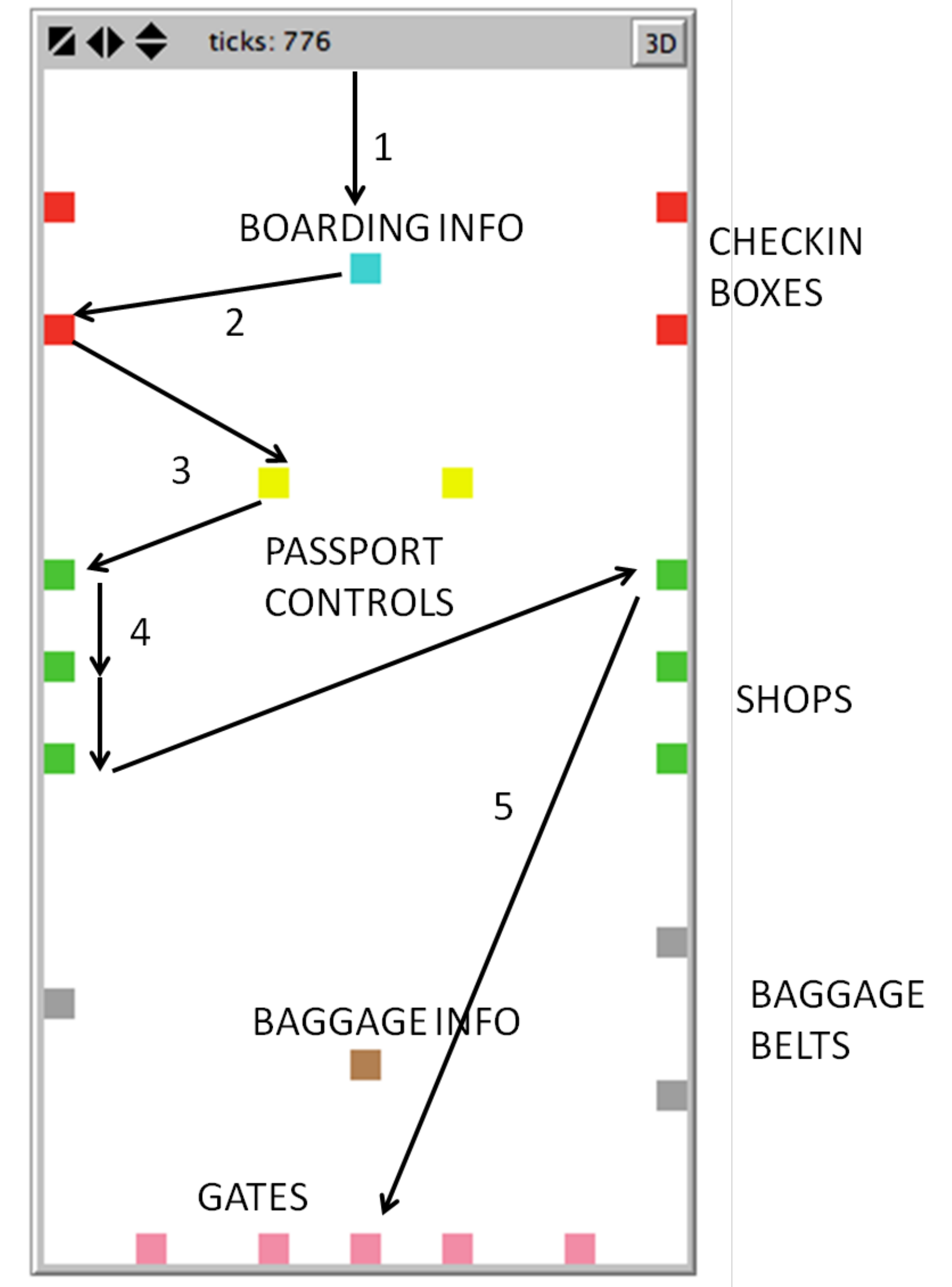}}
 \caption{Followed steps by ingoing agents with AmI}
 \label{fig:perceptionz}
\end{figure}

In this section we define (using NetLogo) a simulated scenario where the architecture described in section 3 MAS for context-aware problems can be applied. This scenario would allow us to consider two evaluation criteria that would become more discriminant when there are many agents in the system: First, satisfaction provided by Ambient Intelligence, which is linked to the accomplishment of agent’s goals through an appropriate use of time. This concept is computed according to three satisfaction evaluation criteria: whether we achieved the main goal (to avoid missing the plane) or not, how much we met desired activities (shopping) and how much we avoided undesired activities (time spent in queues/lines). For instance, an agent nor satisfied at all would have missed the plane, and an agent would be mostly not satisfied if it did not buy any gift according to his shopping interests or if  he spent a lot of time in lines. 
The second is time saving obtained through the use of context information. A correct use of information in our domain stands for avoiding going to the information panels of the airport, and avoiding going around while shopping (through the use of location indications). For instance, an agent would not have saved any time if he had  walked to the flight information panel and to the boarding information panel; furthermore it would have taken a detour (instead following a straight course) to reach the provider that fits its shopping interests. These time savings can be obtained using information provided by AmI. The corresponding difference in the steps followed by ingoing agents with and without AmI in Figures \ref{fig:perceptionzingoingami} and \ref{fig:perceptionzingoing} are:
\begin{enumerate}
\item request the service from Boarding Info.
\item request the service from Checkin Counter.
\item request the service from Passport Control.
\item request the service from Shops (until finding the one that matched with the shopping interest of the agent).
\item request the service from Boarding Gates.
\end{enumerate}

\begin{figure}
 \centering
  \subfloat[Outgoing 1]{
   \label{fig:perceptionzoutgoingami}
    \includegraphics[height=7.5 cm]{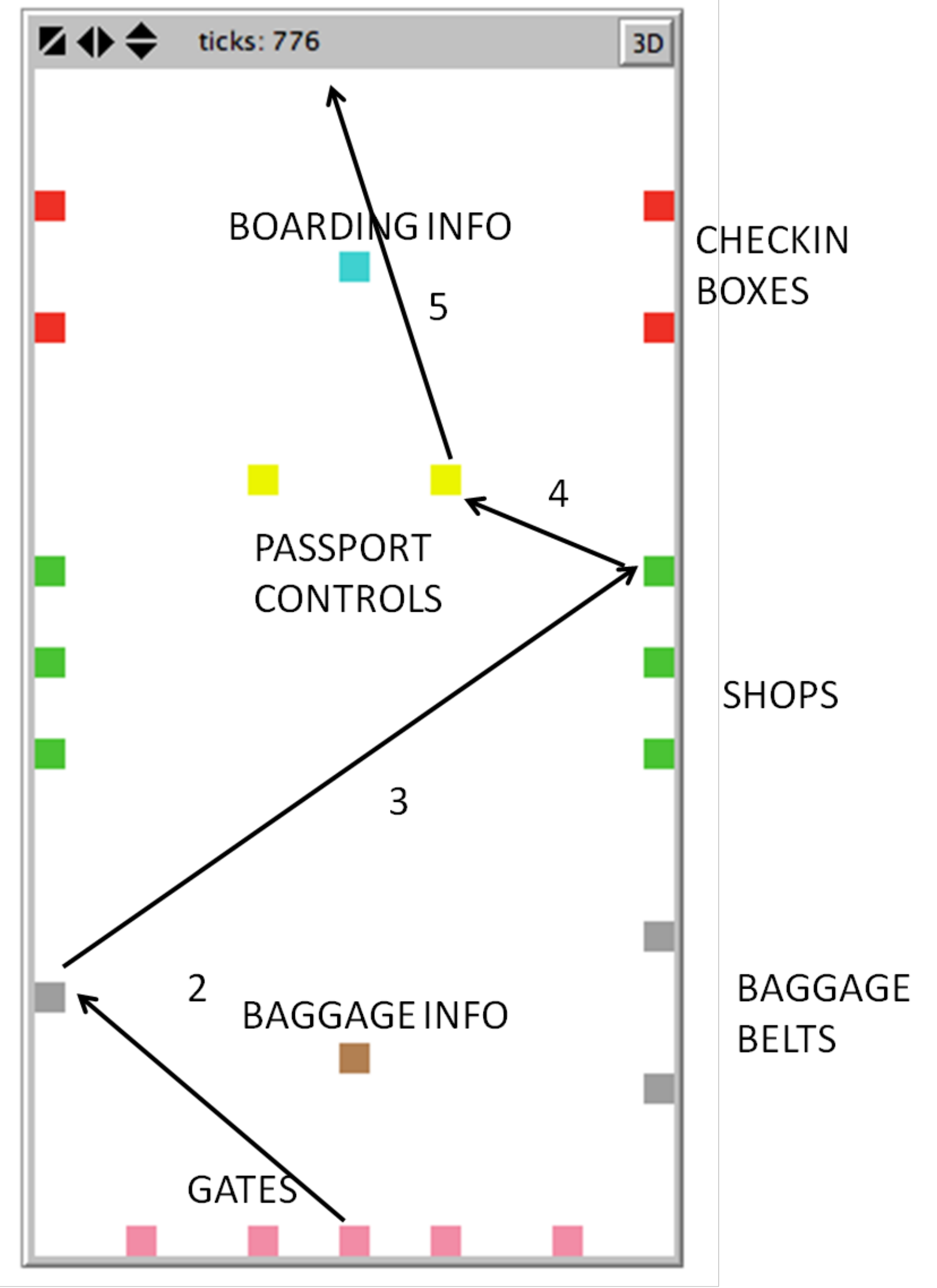}}
  \subfloat[Outgoing 2]{
   \label{fig:perceptionzoutgoing}
    \includegraphics[height=7.5 cm]{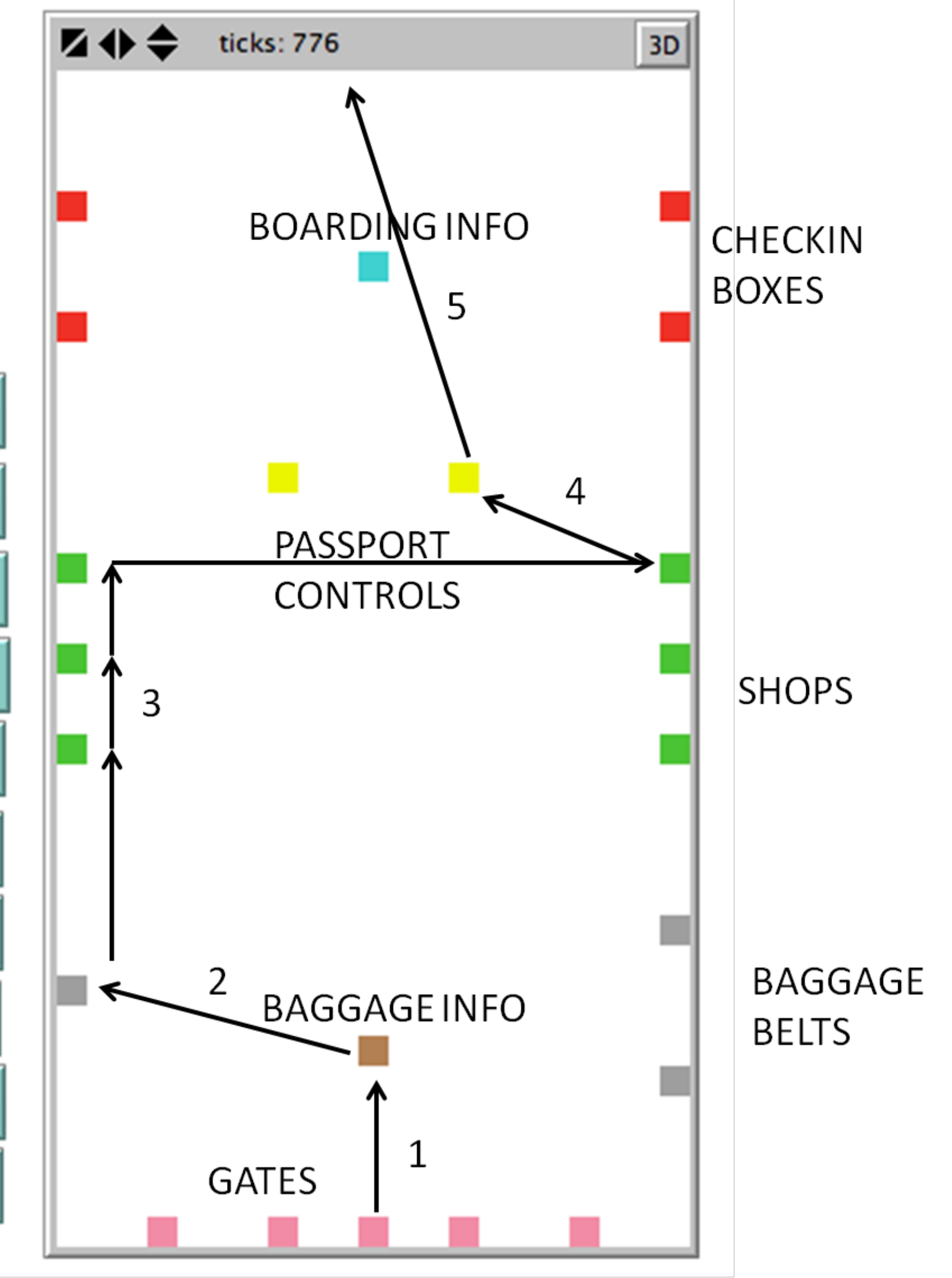}}
 \caption{Followed steps by outgoing agents with AmI}
 \label{fig:perceptionz}
\end{figure}

The corresponding steps followed by outgoing agents with and without AmI in Figure \ref{fig:perceptionzoutgoingami} and \ref{fig:perceptionzoutgoing} are:
\begin{enumerate}
\item request the service from Baggage Info.
\item request the service from Baggage Belt.
\item request the service from Shops (until finding the one that matched with the shopping interest of the agent).
\item request the service from Passport Control.
\item go outside.
\end{enumerate}

Thus, this NetLogo model of an airport includes several types of User and Provider Agents (besides the aforementioned Positioning, Facilitator and Evaluator agents of our MAS architecture for context-aware problems). User agents may be passengers, crew and staff, but additionally they may be of two types (outgoing and ingoing), passing through several services located in a specific order on  a map: 
 
\begin{itemize}
\item Outgoing agents go through main entrance, flight information panel, check-in counter, passport control, shops, boarding information panel and boarding gates.
\item Ingoing agents go through boarding gate, go to baggage information panel, baggage belt, shops, passport control and main entrance. 
\end{itemize}

The eight concepts and six predicates that formed the ontology of the system were used in the FIPA communications in NetLogo. The equivalent OWL ontology can be obtained using the OWL-API (3.1.0) \citep{netlogoowl} that extracts state and structure ontologies from an existing Netlogo model. 
We can observe how the elements of the ontology were used in the next couple of FIPA communications examples of our model:

\begin{verbatim}
(turtle 51): 
["inform" "sender:0" "receiver:51" "content:" "isProvider (Place 
(Building Airport ; Floor 0); Position: Belt (patch 18 6) ; 
AID: 51 ; Service (Name: Baggage-Delivery) )"]

(turtle 2): 
["request" "sender:51" "receiver:2" "content:" "Provide (Product 
(Name: Baggage-Delivery ; Characteristics: Baggage-Number 1 ) ; 
AID: 51 )"]
\end{verbatim}

We assume that each of the user agents has defined a particular predefined profile (traveling profile and personal profile), corresponding to the features of the profile concept of our ontology, which gives values to the following attributes:
\begin{itemize}
\item How much interest the agent has in each type of shop.
\item How much baggage he is carrying (number of suitcases). 
\item How much estimated danger perception may be produced for external observers (due to his physical aspect, nationality, etc.).
\item Flight number. 
\end{itemize}
User Agents go shopping if they have enough (estimated) time to do so. We use randomly generated initial data of passenger profiles, so the model is just an approximation of real-world airports.

The concept Service is instantiated with Airport services that are provided by Agent Providers:
\begin{itemize}
\item Check-in counter.
\item Passport control.
\item Shops.
\item Baggage belt.
\item Boarding gate.
\end{itemize}

Furthermore queues are formed in services (check-in counters, passport controls, shops, baggage belts and boarding gates), and User Agents have to wait until the Agent Provider is not busy. We assume that information panels do not consume time and do not produce any lines. 
In order to evaluate the benefits of using context with our MAS architecture, there will be some User Agents that use AmI and others who do not. Information panels are Facilitator agents for the agents using Am and Provider agents for the agents not using AmI. Each of these agents using AmI would be executing the communications with Positioning and Facilitator agents (included in the 12-steps protocol described in section 3), and we assume that such communications also involve a relatively short elapsed time and also form lines to attend User Agents. But on the other hand, for instance, User Agents using context do not require  passing through information panels, and they know the exact location of the most interesting shops (for that particular agent) thanks to communication with Facilitator agents, avoiding a random walk through the shops that  users  not using AmI have to take. We also assume that moving through the map requires time (agents move 1 position per iteration) and providing services has an estimated time (random distribution of different types that depends on profile features :  more baggage, more time in the check-in counter, more perception of danger, more time in passport control). Since the same instance of our user agents do not repeat model executions, The evaluator agent makes no sense in this simulation. Otherwise, the evaluator agent would allow agents using AmI to know a priori what check-in counter  to use (because the user always travels with the same company) or the boarding gate (because the user always travels to the same destination) or the shops to purchase in (since it knows the shopping preference) without the participation of the Facilitator agent. 
We can observe these differences in the sequence of intentions (coded in reversed order) that the four types of agents execute. For instance, ingoing agents that do not use Ambient Intelligence have to execute intentions for moving to the baggage info screen in order to know the belt number corresponding to their  flight, and move through different shops until they find the most interesting shop they are looking for.

\begin{verbatim}
  add-intention "move-to-output" "in-output"
  add-intention "pass-control" "past-control"
  add-intention "move-to-control" "in-control"
  add-intention "shopping" "shopped"
  add-intention "move-to-shops" "in-shops"
  add-intention "collect-baggage" "baggage-collected"
  add-intention "move-to-belt" "in-belt"
  add-intention "ask-baggage-info" "informed-belt-baggage"
  add-intention "move-to-baggage-info" "in-baggage-info"
\end{verbatim}
On the other hand, agents that use Ambient Intelligence do not require moving to the baggage info screen, and they move directly to the most interesting shop as the next Netlogo code shows:
\begin{verbatim}
  add-intention "move-to-output" "in-output"
  add-intention "pass-control" "past-control"
  add-intention "move-to-control" "in-control"
  add-intention "shopping" "shopped"
  add-intention "move-to-interestingshop" "in-interestingshop";;
  add-intention "collect-baggage" "baggage-collected"
  add-intention "move-to-belt" "in-belt"
  add-intention "ask-baggage-info" "informed-belt-baggage"
\end{verbatim}
Outgoing agents show similar differences according to the use/not use of Ambient Intelligence. Outgoing agents that do not use Ambient Intelligence require Netlogo moving intentions towards check-in and gate info screens in order to know the assigned check-in counter and boarding gates. Additionally these agents would move around shops until they find out the most interesting shop they were looking for.
\begin{verbatim}
  add-intention "move-to-gate" "in-gate"
  add-intention "query-gate" "informed-gate"
  add-intention "move-to-gate-info" "in-gate-info"
  add-intention "shopping" "shopped"
  add-intention "move-to-shops" "in-shops"
  add-intention "pass-control" "past-control"
  add-intention "move-to-control" "in-control"
  add-intention "request-checkin" "done-checkin"
  add-intention "move-to-checkin" "in-checkin"
  add-intention "query-checkin" "informed-checkin"
  add-intention "move-to-checkin-info" "in-checkin-info"
\end{verbatim}
While outgoing agents that use Ambient Intelligence would not require going to the info screens, and they move directly to the most interesting shop as it shows the next code corresponding to their Netlogo intentions to be executed in reversed order:
\begin{verbatim}
  add-intention "move-to-gate" "in-gate"
  add-intention "query-gate" "informed-gate"
  add-intention "shopping" "shopped"
  add-intention "move-to-interestingshop" "in-interestingshop"
  add-intention "pass-control" "past-control"
  add-intention "move-to-control" "in-control"
  add-intention "request-checkin" "done-checkin"
  add-intention "move-to-checkin" "in-checkin"
  add-intention "query-checkin" "informed-checkin"
\end{verbatim}

The definition of this model allows us to simulate several runnings of high- populated agent systems moving from the airport main entrance to boarding gates and the opposite. The first evaluation criteria consists in comparing satisfaction provided by the activities carried out by agents in the airport; although it is subjective we quantified it assigning satisfaction values to the next circumstances as follows: 
\begin{itemize}
\item Avoiding missing the flight (high positive value).
\item Shopping pleasure (low positive value).
\item Time spent in lines (low negative value).
\end{itemize}

On the other hand, the second  criterion is measured with the average time-spent in the airport.

Initial setup parameters of each simulation running are:
\begin{itemize}
\item Number of ingoing agents who do not use AmI.
\item Number of ingoing agents who use AmI.
\item Number of outgoing agents who do not use AmI.
\item Number of outgoing agents who use AmI.
\item Number of iterations required to avoid missing the flight.
\item Number of passport controls.
\item Number of check-in counters.
\item Number of shops of different types.
\item Number of boarding gates.
\item Number of baggage belts.
\end{itemize}

Different values of these initial parameters would setup models of different types of (small and big) airports.

Further details of the implementation can be observed since the code can be downloaded at sourceforge:\\ 
\url{https://sourceforge.net/projects/netlogo-bdi-fipa-airport-model/}. 

Additionally we have already uploaded our NetLogo model into the official NetLogo library at: 

\url{http://ccl.northwestern.edu/netlogo/models/index.cgi}

\subsection{Results}

\begin{figure}
\centering
\includegraphics[height=9 cm]{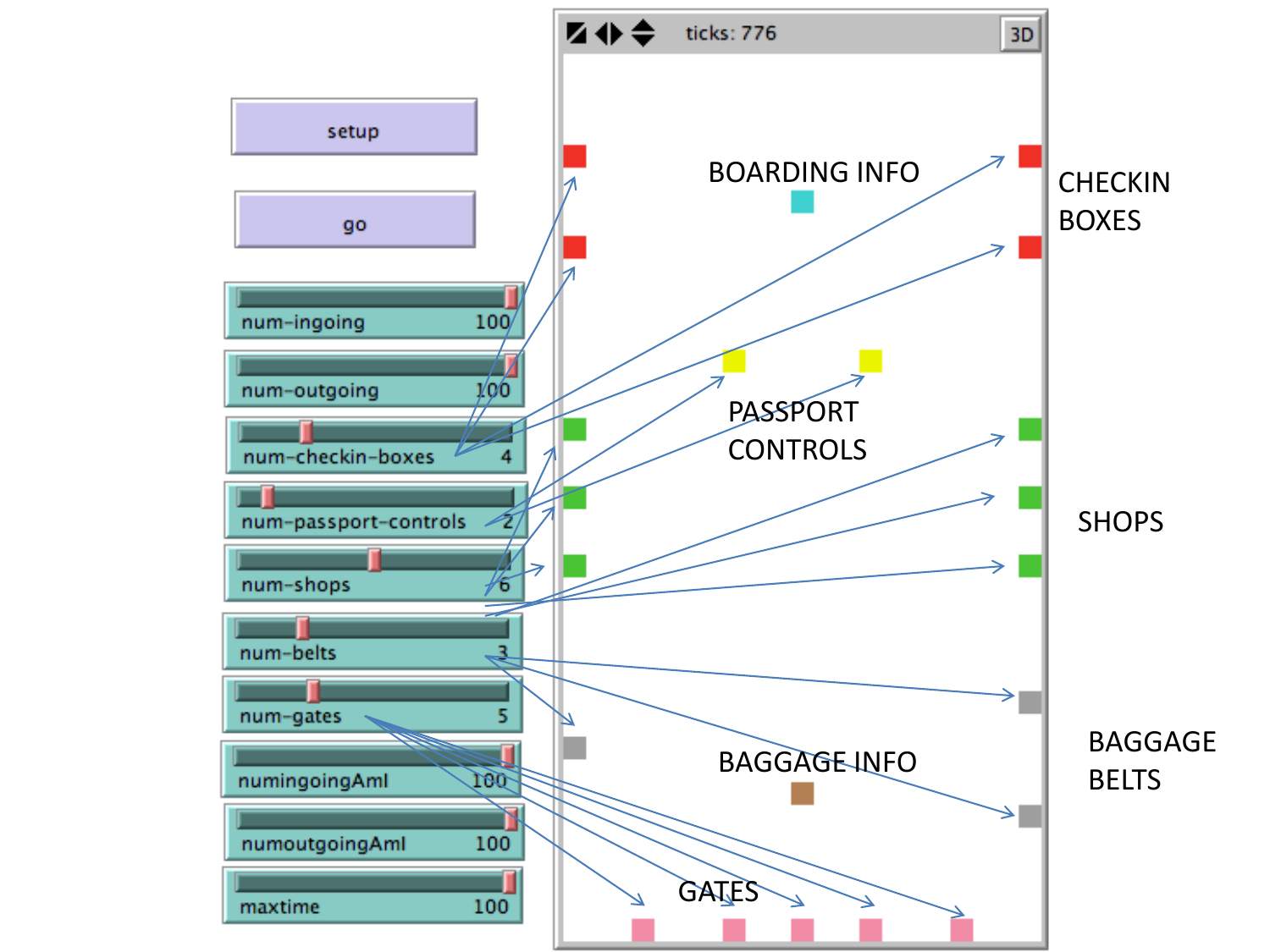}
\caption{Description of elements in our NetLogo Model and initial parameter setup}
\label{fig:perceptionx}
\end{figure}

Figure \ref{fig:perceptionx} shows a caption of the NetLogo agent simulation that describes the elements representing rooms of the airport: where red points represent check-in counters, the cyan point represents flight information panels, the yellow points represent passport controls, the black line represents a wall, the green points represent shops, the brown point represents both boarding information panel and baggage belt information panel, and finally, the pink points represent boarding gates.

\begin{figure}
\centering
\includegraphics[height=8 cm]{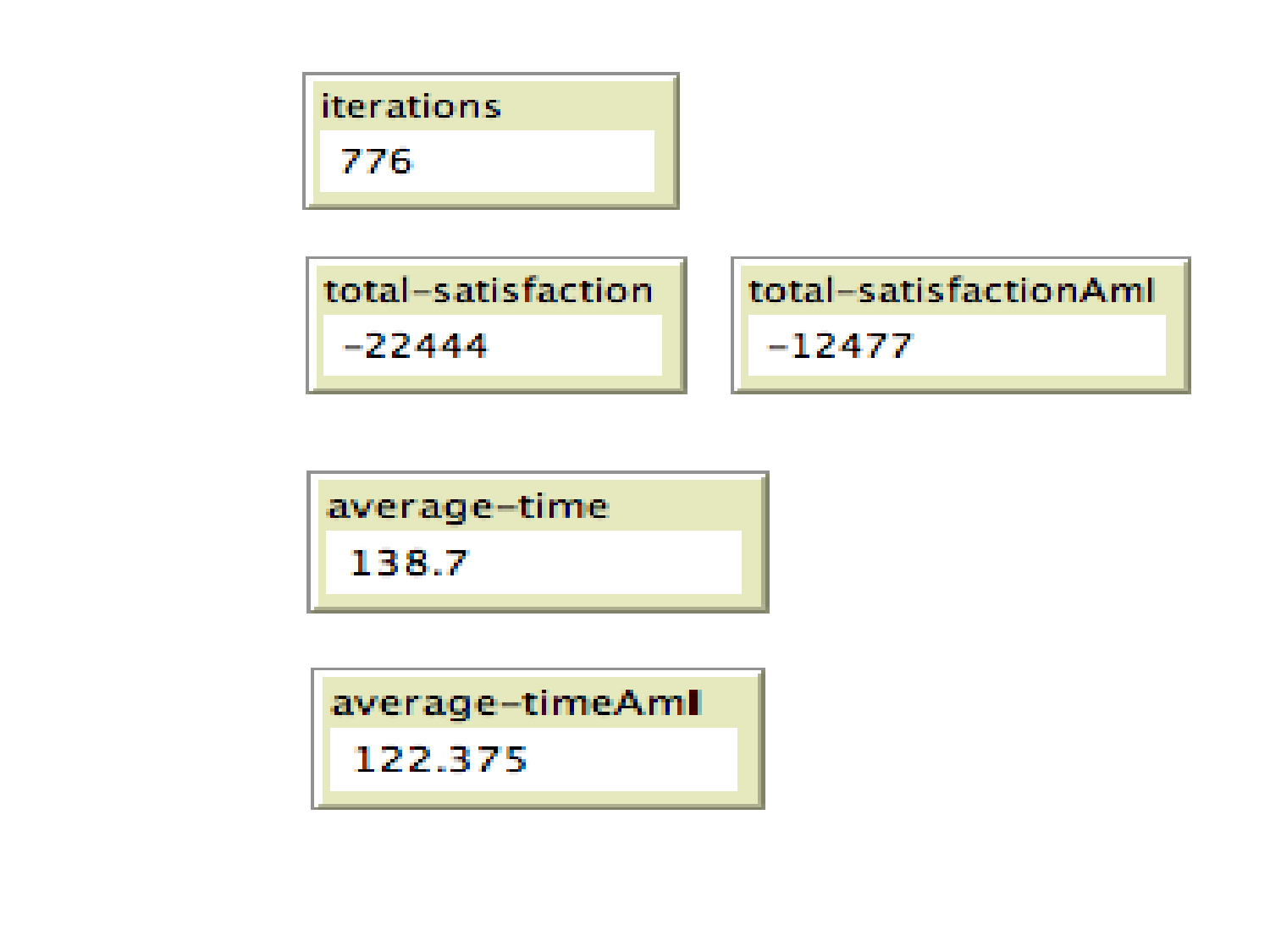}
\caption{Final outcome of a NetLogo simulation of the airpot }
\label{fig:perception3}
\end{figure}

\begin{figure}
\centering
\includegraphics[height= 8 cm]{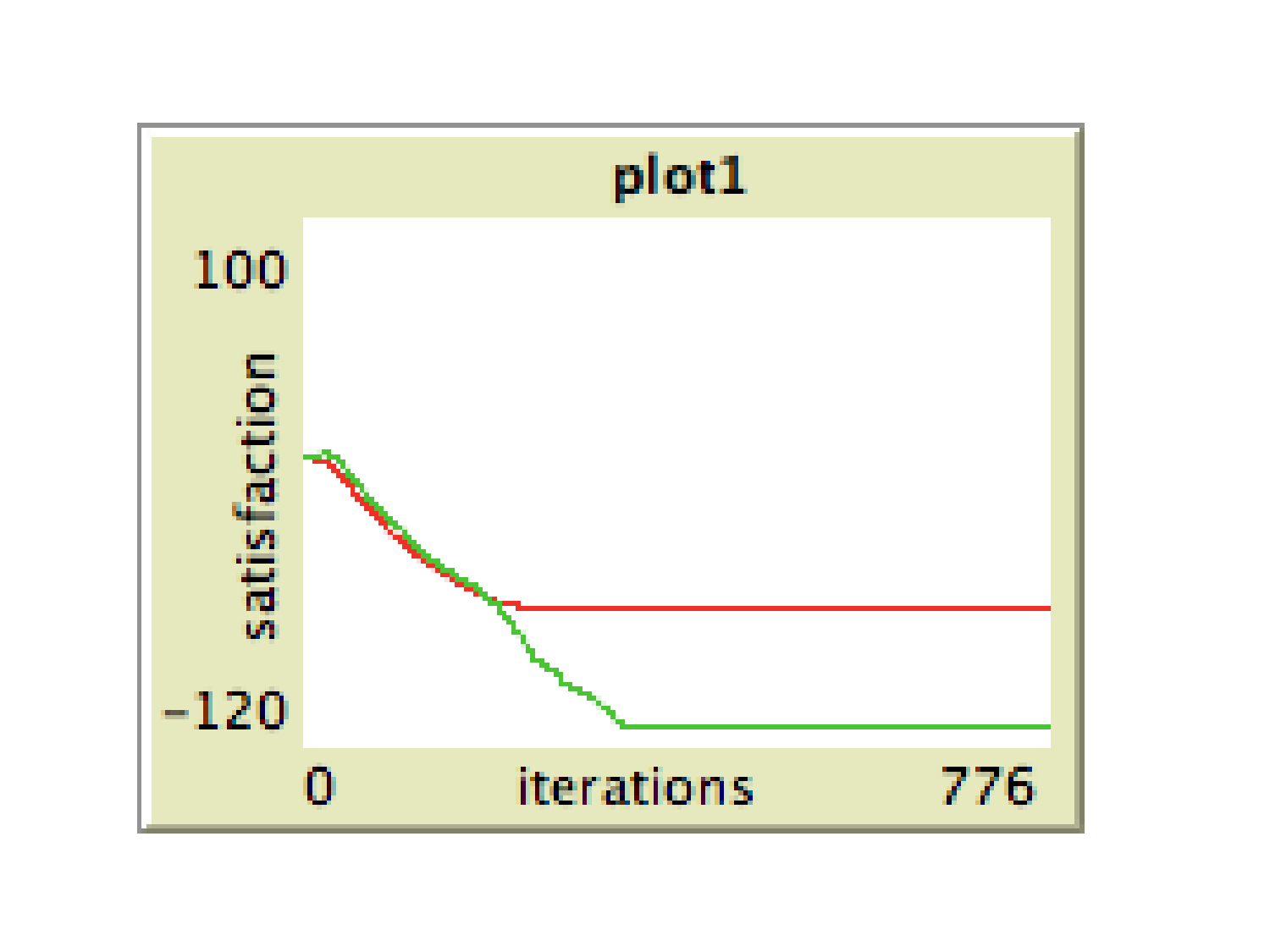}
\caption{Evolution of satisfaction values with and without AmI }
\label{fig:perception4}
\end{figure}

The figure \ref{fig:perceptionx} generated by NetLogo shows the definition of the initial parameter setup that would define the scale of each execution of the model, while 5 shows the relevant output variables that a NetLogo execution shows: total satisfaction of agents not using AmI, total satisfaction of agents using AmI, the average-time spent in the airport of AmI agents and average-time spent in the airport of nonAmI agents. These are the values we were trying to obtain in order to evaluate the benefits of using AmI in a context-aware scenario. Finally, a curve of the evolution of both total satisfaction values can also be observed in 6. These curves show how satisfaction is very similar in the beginning of each simulation run, but as service lines increase and user agents miss their flights, satisfaction  is reduced, but always more so for non AmI agents. Values of average time also worsen when a very high number of agents are included in the simulation. So both evaluation criteria (time savings and agent satisfaction) show the potential benefits of using AmI when there is a sufficiently high number of agents. 
According to http://ccl.northwestern.edu/netlogo/docs/faq.html , the FAQ section of the official website, we have tried 30 runs with the initial parameter setup shown in NetLogo version 5 which runs models in a scientifically reproducible way. Table 1 shows a time-saving improvement of approximately 18\%, and an approximately 40\% improvement in agent satisfaction for agents using AmI. Other simulations with different initial parameter setup fit approximately  these patterns, while the same number of AmI and nonAmi agents (50\% each) participate in the simulation. 
\begin{table} 
\centering
\caption{Results of 30 NetLogo executions of airport model}
\label{table8}
\begin{tabular}{|c|c|c|}
\hline Name & Average & Standard Deviation \\
\hline total-satisfaction & -21858,51 & 1415,67 \\
total-satisfactionAmI & -13596,34 & 1937,87 \\
average-time & 145,31 & 7,26 \\
average-timeAmI & 122,87 & 4,93 \\
\hline
\end{tabular}
\end{table}

\section{Conclusions}
In this contribution we looked forward to estimate the potential benefits of using an AmI application of agents already defined by us in an airport domain. Since we have previously suggested an agent architecture, an ontology and a 12-step protocol to provide AmI services in such a domain, we were interested in transforming such issues into a simulation that could easily visualize and compute such benefits when the number of agents is high enough. Although we had a JADE implementation of this model\citep{sanchez-jade}, we observed that a NetLogo model could achieve these goals. Since our initial proposal included FIPA messages and BDI reasoning agents, we used both NetLogo extensions to satisfy both requirements. We also, for simplicity sake, re-introduce an equivalent of our (small-sized) ontology into NetLogo instead of using an external already defined protege ontology. Although initial data in simulations are generated randomly, and the model is just an approximation of real-world airports, initial parameters allow representation of both small and large airports through different values in  the number of boarding gates, shops, check-in counters, baggage belts, etc. The definition of this case of use opens up an interesting way to evaluate agent approaches dedicated to AmI, which is a significant contribution to the final development of AmI. In spite of the interaction complexity (12 step protocol to provide services in AmI), we use a very limited number of options, so the internal reasoning of agents is very straightforward, which is a limitation imposed by Netlogo simplicity. But this platform allows  testing the consequences of using different interaction protocols when the number of involved agents is high, ignoring or simplifying the computational overhead that BDI reasoning and FIPA protocols impose over other alternative agent implementations. 
Our proposal addresses the three most common shortcomings of AmI simulations according to \citep{mercedesgarijo}: 
\begin{itemize}
\item simulations are closed, and can not be parameterized.
\item experiments are not reproducible
\item source code is rarely given
\end{itemize}

These simulation results help us to establish and quantify the potential benefits of using AmI. It also provides us with an estimate for an  airport scenario where we should put the effort depending on if it is useful to use AmI for in this kind of scenario. Experiments and graphics resulting from computing multiple runs with equal numbers of each agent type could lead  us to conclusions about the possible relevance of using the AmI facilities in this environment. Therefore, through a context-specific model we have measured the benefits of using AmI; this evaluation task is innovative, particularly because we it was done without  oversimplification that would requiere removing BDI or FIPA in our model. As future work we  would like to include experiments with different agent system architectures, and a very different population composition (the proportion of AmI vs. NonAmI agents) for this airport scenario. Additionally, we plan to characterize agents in a richer way by including  such features as: excitement, anxiety, urgency and fatigue together with types of agents different from ingoing/outgoing passengers: staff, tourists, business people, groups, etc. Finally, more complex airport maps based on real airports could increase the realism of the simulations.

\section*{Acknowledgements}

This work was partially funded by CNPq PVE Project 314017/2013-5, FAPERJ APQ1 Project 211.500/2015 and by Projects MINECO TEC2012-37832-C02-01, CICYT TEC2011-28626-C02-02.

\balance
\bibliographystyle{rusnat}
\bibliography{bibs}

\end{document}